\documentclass{IOS-Book-Article}

\usepackage{amssymb,amsmath}

\usepackage{txfonts} 
\usepackage[scaled=0.85]{helvet} 
\usepackage[T1]{fontenc}

\usepackage{url}
\usepackage[bookmarks=false,pdfpagelabels,bookmarksnumbered,bookmarksopen,bookmarksopenlevel=1,colorlinks=true,
citecolor=black, filecolor=black, linkcolor=black, menucolor=black,
urlcolor=black, pdftitle={A Description Logic Primer},
pdfauthor={Markus Kroetzsch, Frantisek Simancik, Ian Horrocks}, pdfkeywords={description logics, introduction, SROIQ, syntax, semantics}]{hyperref}
\usepackage{colortbl}
\usepackage[table]{xcolor}

\newcommand{\concept}[1]{\mathsf{#1}}
\newcommand{\role}[1]{\mathsf{#1}}
\newcommand{\individual}[1]{\mathsf{#1}}

\newcommand{\axiom}{\alpha}
\newcommand{\ontology}{\mathcal{O}}
\newcommand{\I}{\mathcal{I}}

\newcommand{\atleast}[1]{\mathord{\geqslant}#1\,}
\newcommand{\atmost}[1]{\mathord{\leqslant}#1\,}
\newcommand{\Self}{\mathit{Self}}
\newcommand{\self}[1]{\exists{#1}.\Self}

\newcommand{\disjointop}{\mathit{Disjoint}}
\newcommand{\disjoint}[2]{\disjointop(#1, #2)}

\newcommand{\set}[1]{\{{#1}\}}
\newcommand{\tuple}[1]{\langle{#1}\rangle}

\newcommand{\rolnames}{{\text{\sf{N}}_R}} 
\newcommand{\connames}{{\text{\sf{N}}_C}} 
\newcommand{\indnames}{{\text{\sf{N}}_I}} 
\newcommand{\concepts}{\mathbf{C}} 
\newcommand{\roles}{\mathbf{R}} 

\newcommand{\SROIQ}{\mathcal{SROIQ}}

\makeatletter
\def\title@note@fmt{\def\thefootnote{\fnsymbol{footnote}}}
\makeatother

\begin{document}

\begin{frontmatter}                           

\title{A Description Logic Primer\thanks{Version 1.2 of 3 June 2013.
Comments and suggestions can be sent to Markus Kr\"{o}tzsch at markus.kroetzsch@cs.ox.ac.uk. This document can freely be used and distributed under the terms of CC \href{http://creativecommons.org/licenses/by-nc-sa/3.0/}{By-SA-NC 3.0}. Please contact the authors if you would like to reproduce this document under another license.}}


\author{\href{http://korrekt.org/}{Markus Kr\"{o}tzsch}},
\author{\href{http://www.cs.ox.ac.uk/isg/people/frantisek.simancik/}{Franti\v{s}ek Siman\v{c}\'{i}k}},
\author{\href{http://www.cs.ox.ac.uk/ian.horrocks/}{Ian Horrocks}}%

\address{Department of Computer Science, University of Oxford, UK}

\begin{abstract}
This paper provides a self-contained first introduction to description logics (DLs). The main concepts and features are explained with examples before syntax and semantics of the DL $\SROIQ$ are defined in detail. Additional sections review lightweight DL languages, discuss the relationship to the OWL Web Ontology Language and give pointers to further reading.
\end{abstract}

\end{frontmatter}

\setcounter{footnote}{0}

\section*{Introduction}

Description logics (DLs) are a family of knowledge representation languages that are widely used in ontological modelling. An important practical reason for this is that they provide one of the main underpinnings for the OWL Web Ontology Language as standardised by the World Wide Web Consortium (W3C). However, DLs have been used in knowledge representation long before the advent of ontological modelling in the context of the Semantic Web, tracing back to first DL modelling languages in the mid 1980s.

As their name suggests, DLs are logics (in fact most DLs are decidable fragments of first-order logic), and as such they are equipped with a \emph{formal semantics}: a precise specification of the meaning of DL ontologies. This formal semantics allows humans and computer systems to exchange DL ontologies without ambiguity as to their meaning, and also makes it possible to use logical deduction to \emph{infer} additional information from the facts stated explicitly in an ontology -- an important feature that distinguishes DLs from other modelling languages such as UML.

The capability of inferring additional knowledge increases the modelling power of DLs but it also requires some understanding on the side of the modeller and, above all, good tool support for computing the conclusions. The computation of inferences is called \emph{reasoning} and an important goal of DL language design has been to ensure that reasoning algorithms of good performance are available. This is one of the reasons why there is not just a single description logic: the best balance between expressivity of the language and complexity of reasoning depends on the intended application.

In this paper we provide a self-contained first introduction to description logics. We start by explaining the basic way in which knowledge is modelled in DLs in Section~\ref{sec:basic} and continue with an intuitive introduction to the most important DL modelling features in Section~\ref{sec:construct}. This leads us to the rather expressive DL called $\SROIQ$, the syntax of which we summarise in Section~\ref{sec:syntax}. In Section~\ref{sec:semantics}, we explain the underlying ideas of DL semantics and use it to define the meaning of $\SROIQ$ ontologies. Many DLs can be obtained by omitting some features of $\SROIQ$ and in Section~\ref{sec:fragments} we review some of the most important DLs obtained in this way. In particular, this includes various lightweight description logics that allow for particularly efficient reasoning. In Section~\ref{sec:owl} we discuss the relationship of DLs to the OWL Web Ontology Language. We conclude with pointers to further reading in Section~\ref{sec:reading}.

\section{Basic Building Blocks of DL Ontologies}\label{sec:basic}

Description logics (DLs) provide means to model the relationships between entities in a domain of interest. In DLs there are three kinds of entities: concepts, roles and individual names.\footnote{In OWL concepts and roles are respectively known as classes and properties; see Section~\ref{sec:owl}.} Concepts represent sets of individuals, roles represent binary relations between the individuals, and individual names represent single individuals in the domain. Readers familiar with first-order logic will recognise these as unary predicates, binary predicates and constants.

For example, an ontology modelling the domain of people and their family relationships might use concepts such $\concept{Parent}$ to represent the set of all parents and $\concept{Female}$ to represent the set of all female individuals, roles such as $\role{parentOf}$ to represent the (binary) relationship between parents and their children, and individual names
such as $\individual{julia}$ and $\individual{john}$ to represent the individuals Julia and John.

Unlike a database, a DL ontology does not fully describe a particular situation or ``state of the world''; rather it consists of a set of statements, called axioms, each of which must be true in the situation described. These axioms typically capture only partial knowledge about the situation that the ontology is describing, and there may be many different states of the world that are consistent with the ontology.
Although, from the point of view of logic, there is no principal difference between different types of axioms, it is customary to separate them into three groups: assertional (ABox) axioms, terminological (TBox) axioms and relational (RBox) axioms.

\subsection{Asserting Facts with ABox Axioms}
\label{subsection:abox}

ABox axioms capture knowledge about named individuals, i.e., the concepts to which they belong and how they are related to each other. The most common ABox axioms are \emph{concept assertions} such as
\begin{equation}
\label{mother:julia}
\concept{Mother}(\individual{julia}),
\end{equation}
which asserts that Julia is a mother or, more precisely, that the individual named $\individual{julia}$ is an \emph{instance} of the concept $\concept{Mother}$.

\emph{Role assertions} describe relations between named individuals. The assertion
\begin{equation}
\label{parentOf:julia:john}
\role{parentOf}(\individual{julia},\individual{john}),
\end{equation}
for example, states that Julia is a parent of John or, more precisely, that the individual named $\individual{julia}$ is in the relation that is represented by $\role{parentOf}$ to the individual named $\individual{john}$. The previous sentence shows that it can be rather cumbersome to explicitly point out that the relationships expressed by an axiom are really relationships between the individuals, sets and relations that are represented by the respective individual names, concepts and roles. Assuming that this subtle distinction between syntactic identifiers and semantic entities is understood, we will thus often adopt a more sloppy and readable formulation. Section~\ref{sec:semantics} below explains the underlying semantics with greater precision.

Although it is intuitively clear that Julia and John are different individuals, this fact does not logically follow from what we have stated so far. DLs do not make the \emph{unique name assumption}, so different names might refer to the same individual unless explicitly stated otherwise. The \emph{individual inequality} assertion
\begin{equation}
\label{julia:neq:john}
\individual{julia} \not\approx \individual{john}
\end{equation}
is used to assert that Julia and John are actually different individuals.
On the other hand, an \emph{individual equality} assertion, such as
\begin{equation}
\label{john:eq:johnny}
\individual{john} \approx \individual{johnny},
\end{equation}
states that two different names are known to refer to the same individual. Such situations can arise, for example, when combining knowledge about the same domain from several different sources, a task that is known as \emph{ontology alignment}.

\subsection{Expressing Terminological Knowledge with TBox Axioms}
\label{subsection:tbox}

TBox axioms describe relationships between concepts. For example, the fact that all mothers are parents is expressed by
the \emph{concept inclusion}
\begin{equation}
\label{mother:sub:parent}
\concept{Mother} \sqsubseteq \concept{Parent},
\end{equation}
in which case we say that the concept $\concept{Mother}$ is \emph{subsumed} by the concept $\concept{Parent}$. Such knowledge can be used to infer further facts about individuals. For example, \eqref{mother:julia} and \eqref{mother:sub:parent} together imply that Julia is a parent.

\emph{Concept equivalence} asserts that two concepts have the same instances, as in
\begin{equation}
\label{person:eq:human}
\concept{Person} \equiv \concept{Human}.
\end{equation}
While synonyms are an obvious example of equivalent concepts, in practice one more often uses concept equivalence to give a name to complex expressions as introduced in Section~\ref{subsection:boolean:constructors} below.
Furthermore, such additional concept expressions can be combined with equivalence and inclusion to describe more complex situations such as the disjointness of concepts, which asserts that two concepts do not share any instances.

\subsection{Modelling Relationships between Roles with RBox Axioms}
\label{subsection:rbox}

RBox axioms refer to properties of roles. As for concepts, DLs support \emph{role inclusion} and \emph{role equivalence} axioms. For example, the inclusion
\begin{equation}
\label{parentOf:sub:ancestorOf}
\role{parentOf} \sqsubseteq \role{ancestorOf}
\end{equation}
states that $\role{parentOf}$ is a \emph{subrole} of $\role{ancestorOf}$, i.e., every pair of individuals related by $\role{parentOf}$ is also related by $\role{ancestorOf}$. Thus \eqref{parentOf:julia:john} and \eqref{parentOf:sub:ancestorOf} together imply that Julia is an ancestor of John.

In role inclusion axioms, \emph{role composition} can be used to describe roles such as
$\role{uncleOf}$. Intuitively, if Charles is a brother of Julia and Julia is a
parent of John, then Charles is an uncle of John. This kind of relationship
between the roles $\role{brotherOf}$, $\role{parentOf}$ and $\role{uncleOf}$ is
captured by the \emph{complex role inclusion} axiom
\begin{equation}
\label{brotherOf:comp:parentOf:sub:uncleOf}
\role{brotherOf} \circ \role{parentOf} \sqsubseteq \role{uncleOf}.
\end{equation}
Note that role composition can only appear on the left-hand side of complex role inclusions. Furthermore, in order to retain decidability of reasoning (see the end of Section~\ref{sec:semantics} for a discussion on decidability), complex role inclusions are governed by additional structural restrictions that
specify whether or not a collection of such axioms can be used together in one ontology.

Nobody can be both a parent and a child of the same individual, so the two roles $\role{parentOf}$ and $\role{childOf}$ are disjoint. In DLs we can write \emph{disjoint roles} as follows:
\begin{equation}
\disjoint{\role{parentOf}}{\role{childOf}}.
\end{equation}

Further RBox axioms include \emph{role characteristics} such
as reflexivity, symmetry and transitivity of roles. These are closely related to a number of other DL features and we
will discuss them again in more detail in Section~\ref{subsection:role:characteristics}.

\section{Constructors for Concepts and Roles}\label{sec:construct}

The basic types of axioms introduced in Section~\ref{sec:basic} are rather limited for accurate modelling. To describe more complex situations, DLs allow new concepts and roles to be built using a variety of different constructors. We distinguish concept and role constructors depending on whether concept or role expressions are constructed. In the case of concepts, one can further separate basic Boolean constructors, role restrictions and nominals/enumerations. At the end of this section, we revisit the additional kinds of RBox axioms that have been omitted in Section~\ref{subsection:rbox}.

\subsection{Boolean Concept Constructors}
\label{subsection:boolean:constructors}

Boolean concept constructors provide basic Boolean operations that are closely related to the familiar operations of intersection, union and complement of sets, or to conjunction, disjunction and negation of logical expressions.

For example, concept inclusions allow us to state that all mothers are female and that all mothers are parents, but what we really
mean is that mothers are \emph{exactly} the female parents. DLs support such statements by allowing us to form complex concepts such as the \emph{intersection} (also called \emph{conjunction})
\begin{equation}
\label{female:and:parent}
\concept{Female} \sqcap \concept{Parent},
\end{equation}
which represents the set of individuals that are both female and parents. A complex concept can be used in axioms in exactly
the same way as an atomic concept, e.g., in the equivalence $\concept{Mother} \equiv \concept{Female} \sqcap \concept{Parent}$.

\emph{Union} (also called \emph{disjunction}) is the dual of intersection. For example, the concept
\begin{equation}
\label{father:or:mother}
\concept{Father} \sqcup \concept{Mother}
\end{equation}
describes those individuals that are either fathers or mothers. Again, it can be used in an axiom such as $\concept{Parent} \equiv \concept{Father} \sqcup \concept{Mother}$, which states that a parent is either a father or a mother (and vice versa).

Sometimes we are interested in individuals that do \emph{not} belong to a certain concept, e.g., in women who are not married. These could be described by the complex concept
\begin{equation}
\label{female:and:not:married}
\concept{Female} \sqcap \neg\concept{Married},
\end{equation}
where the \emph{complement} (also called \emph{negation}) $\neg \concept{Married}$ represents the set of all individuals that are not married.

It is sometimes useful to be able to make a statement about every individual, e.g., to say that everybody is either male or female. This can be accomplished by the axiom
\begin{equation}
\label{top:sub:male:or:female}
\top \sqsubseteq \concept{Male} \sqcup \concept{Female},
\end{equation}
where the \emph{top concept} $\top$ is a special concept with every individual as an instance; it can be viewed as an abbreviation for $C\sqcup\neg C$ for an arbitrary concept $C$. Note that this modelling is rather coarse as it presupposes that every individual has a gender, which may not be reasonable for instances of a concept such as $\concept{Computer}$. We will see more useful applications for $\top$ later on.

To express that, for the purposes of our modelling, nobody can be both a male and a female at the same time, we can declare the set of male and the set of female individuals to be disjoint. While ontology languages like OWL provide a basic constructor for disjointness, it is naturally captured in DLs with the axiom
\begin{equation}
\label{male:and:female:sub:bottom}
\concept{Male} \sqcap \concept{Female} \sqsubseteq \bot,
\end{equation}
where the \emph{bottom concept} $\bot$ is the dual of $\top$, that is the special concept with no individuals as instances; it can be seen as an abbreviation for $C \sqcap \neg C$ for an arbitrary concept $C$.
The above axiom thus says that the intersection of the two concepts is empty.

\subsection{Role Restrictions}
\label{subsection:role:restrictions}

So far we have seen how to use TBox and RBox axioms to express
relationships between concepts and roles, respectively. The most interesting
feature of DLs, however, is their ability to form statements that link concepts and
roles together. For example, there is an obvious relationship between the concept $\concept{Parent}$ and the role $\role{parentOf}$, namely, a parent is someone who is a parent of at least one individual. In DLs, this
relationship can be captured by the concept equivalence
\begin{equation}
\label{exist:parentOf:top}
\concept{Parent} \equiv \exists \role{parentOf}.\top,
\end{equation}
where the \emph{existential restriction} $\exists \role{parentOf}.\top$ is a
complex concept that describes the set of individuals that are parents of
at least one individual (instance of $\top$). Similarly, the concept $\exists
\role{parentOf}.\concept{Female}$ describes those individuals that are parents
of at least one female individual, i.e., those that have a daughter.

To represent the set of individuals all of whose children are female, we use the \emph{universal restriction}
\begin{equation}
\label{forall:parentOf:female}
\forall \role{parentOf}.\concept{Female}.
\end{equation}
It is a common error to forget that \eqref{forall:parentOf:female} also
includes those individuals that have no children at all. More accurately (and less naturally), the axiom can be said to describe the set of all individuals that have ``no children other than female ones,'' i.e., that have ``no children that are not female.'' Following this wording, the concept \eqref{forall:parentOf:female} could indeed be equivalently expressed as $\neg\exists\role{parentOf}.\neg\concept{Female}$. If this meaning is not intended, one can describe the individuals who have at least one child and with all their children being female by the concept $(\exists \role{parentOf}.\top) \sqcap (\forall \role{parentOf}.\concept{Female})$.

Existential and universal restrictions are useful in combination with
the top concept for expressing \emph{domain} and \emph{range restrictions} on roles; that is, restrictions on the kinds of individual that can be in the domain and range of a given role. To
restrict the domain of $\role{sonOf}$ to male individuals we can use the axiom
\begin{equation}
\label{domain:sonOf:male}
\exists \role{sonOf}.\top \sqsubseteq \concept{Male},
\end{equation}
and to restrict its range to parents we can write
\begin{equation}
\label{range:sonOf:parent}
\top \sqsubseteq \forall\role{sonOf}.\concept{Parent}.
\end{equation}
In combination with the assertion $\role{sonOf}(\individual{john}, \individual{julia})$,
these axioms would then allow us to deduce that John is male and Julia is a parent.
It is interesting to note how this behaviour contrasts with the meaning of \emph{constraints} in databases.
Constraints would also allow us to state, e.g., that all sons must be male.
However, given only the fact that John is a son of Julia, such a constraint would simply be violated (leading to an error) rather than implying that John is male.
Mistaking DL axioms for constraints is a very common source of modelling errors.

\emph{Number restrictions} allow us to restrict the number of individuals that can be reached via a given role. For example, we can form the \emph{at-least restriction}
\begin{equation}
\atleast{2}\role{childOf}.\concept{Parent}
\end{equation}
to describe the set of individuals that are children of at least two
parents, and the \emph{at-most restriction}
\begin{equation}
\atmost{2}\role{childOf}.\concept{Parent}
\end{equation}
for those that are children of at most two parents. The axiom
$\concept{Person} \sqsubseteq {\atleast{2}\role{childOf}.\concept{Parent}}$ $\sqcap\; {\atmost{2}
\role{childOf}.\concept{Parent}}$ then states that every person is a child of
exactly two parents.

Finally, \emph{local reflexivity} can be used to describe the set of individuals that are related to themselves via a given role.
For example, the set of individuals that talk to themselves is described by the concept
\begin{equation}
\self{\role{talksTo}}.
\end{equation}

\subsection{Nominals}
\label{subsection:nominals}

As well as defining concepts in terms of other concepts (and roles), it may also be useful to define a concept by simply enumerating its instances. For example, we might define the concept $\concept{Beatle}$ by enumerating its instances: $\individual{john}$, $\individual{paul}$,
$\individual{george}$, and $\individual{ringo}$.
Enumerations are not supported natively in DLs, but they can
be simulated in DLs using \emph{nominals}. A nominal is a concept that has exactly one instance. For example, $\set{\individual{john}}$ is the concept whose only instance is (the individual represented by) $\individual{john}$.
Combining nominals with union, the
enumeration in our example could be expressed as
\begin{equation}
\label{nominals}
\concept{Beatle} \equiv \set{\individual{john}} \sqcup
\set{\individual{paul}} \sqcup \set{\individual{george}} \sqcup
\set{\individual{ringo}}.
\end{equation}

It is interesting to note that, using nominals, a concept assertion $\concept{Mother}(\individual{julia})$
can be turned into a concept inclusion $\set{\individual{julia}} \sqsubseteq \concept{Mother}$
and a role assertion $\role{parentOf}(\individual{julia}, \individual{john})$
into a concept inclusion $\set{\individual{julia}} \sqsubseteq \exists
\role{parentOf}.\set{\individual{john}}$. This illustrates that the distinction between ABox and TBox does not have a deeper logical meaning.

\subsection{Role Constructors}
\label{subsection:role:constructors}

In contrast to the variety of concept constructors, DLs provide only few
constructors for forming complex roles. In practice, \emph{inverse roles} are the most important such constructor. Intuitively, the relationship between the
roles $\role{parentOf}$ and $\role{childOf}$ is that, for example, if Julia is a
parent of John, then John is a child of Julia and vice versa. More formally,
$\role{parenfOf}$ is the inverse of $\role{childOf}$, which in DLs can be
expressed by the equivalence
\begin{equation}
\role{parentOf} \equiv \role{childOf}^-,
\end{equation}
where the complex role $\role{childOf}^-$ represents the inverse of
$\role{childOf}$.

In analogy to the top concept, DLs also provide the \emph{universal role},
represented by $U$, which always relates all pairs of individuals. It typically plays a minor role in modelling,\footnote{Although there are a few interesting things that could be expressed with $U$, such as \emph{concept products} \cite{RKH-elephants-08}, tool support is rarely sufficient for using this feature in practice.} but it establishes symmetry between roles and concepts w.r.t.\ a top element. Similarly, an \emph{empty role} that corresponds to the bottom concept is also available in OWL but has rarely been introduced as a constructor in DLs; however, we can define any role $R$ to be empty using the axiom $\top\sqsubseteq\neg\exists R.\top$ (``all things do not relate to anything through $R$'').
Interestingly, the universal role cannot be defined by TBox axioms using the constructors introduced above, and in particular universal role restrictions cannot express that a role is universal.

\subsection{More RBox Axioms: Role Characteristics}
\label{subsection:role:characteristics}

In Section~\ref{subsection:rbox} we introduced three forms of RBox axioms: role inclusions, role
equivalences and role disjointness. OWL provides a variety of others, namely role transitivity, symmetry, asymmetry, reflexivity and
irreflexivity. These are sometimes considered as basic axiom types in DLs as well, using some suggestive notation such as $\mathit{Trans}(\role{ancestorOf})$ to express that the role $\role{ancestorOf}$ is transitive. However, such axioms are just syntactic sugar; all role characteristics can be expressed using the features of DLs that we have already introduced.

\emph{Transitivity} is a special form of complex role inclusion. For example, transitivity of $\role{ancestorOf}$ can be captured by the axiom $\role{ancestorOf} \circ \role{ancestorOf} \sqsubseteq \role{ancestorOf}$.  A role is \emph{symmetric} if it is equivalent to its own inverse, e.g., $\role{marriedTo} \equiv \role{marriedTo}^-$, and it is \emph{asymmetric} if it is disjoint from its own inverse, as in $\disjoint{\role{parentOf}}{\role{parentOf}^-}$.
If desired, \emph{global reflexivity} can be expressed by imposing local reflexivity on the top concept as in $\top \sqsubseteq \self{\role{knows}}$. A role is \emph{irreflexive} if it is never locally reflexive, as in the case of $\top \sqsubseteq \neg\self{\role{marriedTo}}$.

\section{The Description Logic $\SROIQ$}\label{sec:syntax}

In this section, we summarise the various features that have been introduced informally above to provide a comprehensive definition of DL syntax. Doing so yields the description logic called $\SROIQ$, which is one of the most expressive DLs commonly considered today. It also largely agrees in expressivity with the ontology language OWL~2 DL, though there are still some differences as explained in Section~\ref{sec:owl}.

Formally, every DL ontology is based on three finite sets of signature symbols: a set $\indnames$ of \emph{individual names}, a set $\connames$ of \emph{concept names} and a set $\rolnames$ of \emph{role names}. Usually these sets are assumed to be fixed for some application and are therefore not mentioned explicitly. Now the set of $\SROIQ$ \emph{role expressions} $\roles$ (over this signature) is defined by the following grammar:
\begin{equation*}
\roles\Coloneqq U\mid \rolnames\mid\rolnames^-
\end{equation*}
where $U$ is the universal role (Section~\ref{subsection:role:constructors}). Based on this, the set of $\SROIQ$ \emph{concept expressions} $\concepts$ is defined as:
\newcommand{\smallmid}{\mid}
\begin{equation*}
\concepts\Coloneqq \connames \smallmid (\concepts\sqcap\concepts) \smallmid (\concepts\sqcup\concepts)  \smallmid \neg\concepts \smallmid \top\smallmid \bot \smallmid \exists\roles.\!\concepts  \smallmid \forall\roles.\!\concepts \smallmid \atleast{n}\roles.\!\concepts \smallmid \atmost{n}\roles.\!\concepts \smallmid \self{\roles} \smallmid \{\indnames\}
\end{equation*}
where $n$ is a non-negative integer. As usual, expressions like $(\concepts\sqcap\concepts)$ represent any expression of the form $(C\sqcap D)$ with $C,D\in\concepts$. It is common to omit parentheses if this cannot lead to confusion with expressions of different semantics. For example, parentheses do not matter for $A\sqcup B\sqcup C$ whereas the expressions $A\sqcap B\sqcup C$ and $\exists R.A\sqcap B$ are ambiguous.

Using the above sets of individual names, roles and concepts, the \emph{axioms} of $\SROIQ$ can be defined to be of the following basic forms:
\begin{align*}
\text{ABox:} & \qquad\concepts(\indnames) \qquad\roles(\indnames,\indnames)\qquad \indnames\approx\indnames\qquad \indnames\not\approx\indnames\\
\text{TBox:} & \qquad\concepts\sqsubseteq\concepts \qquad\concepts\equiv\concepts\\
\text{RBox:} & \qquad\roles\sqsubseteq\roles \qquad\roles\equiv\roles\qquad \roles\circ\roles\sqsubseteq\roles\qquad \disjoint{\roles}{\roles}
\end{align*}
with the intuitive meanings as explained in Section~\ref{sec:basic} and \ref{sec:construct}.

Roughly speaking, a $\SROIQ$ ontology (or \emph{knowledge base}) is simply a set of such axioms. To ensure the existence of reasoning algorithms that are correct and terminating, however, additional syntactic restrictions must be imposed on ontologies. These restrictions refer not to single axioms but to the structure of the ontology as a whole, hence they are called \emph{structural restrictions}. The two such conditions relevant for $\SROIQ$ are based on the notions of \emph{simplicity} and \emph{regularity}. Notably, both are automatically satisfied for ontologies that do not contain complex role inclusion axioms.

A role $R$ in an ontology $\ontology$ is called \emph{non-simple} if some complex role inclusion axiom (i.e., one that uses role composition $\circ$) in $\ontology$ implies instances of $R$; otherwise it is called \emph{simple}.
A more precise definition of the non-simple role expressions of the ontology $\ontology$ is given by the following rules:
\begin{itemize}
\item if $\ontology$ contains an axiom $S\circ T\sqsubseteq R$, then $R$ is non-simple,
\item if $R$ is non-simple, then its inverse $R^-$ is also non-simple,\footnote{If $R=S^-$ already is an inverse role, then $R^-$ should be read as $S$. We do not allow expressions like $S^{- -}$.}
\item if $R$ is non-simple and $\ontology$ contains any of the axioms $R\sqsubseteq S$, $S\equiv R$ or $R\equiv S$, then $S$ is also non-simple.
\end{itemize}
All other roles are called simple.\footnote{Whether the universal role $U$ is simple or not is a matter of preference that does not affect the computational properties of the logic \cite{RKH:Jelia-08}. However, the universal role in OWL~2 is considered non-simple.} Now for a $\SROIQ$ ontology it is required that the following axioms and concepts contain simple roles only:
\begin{align*}
\text{Restricted axioms:} & \qquad\disjoint{\roles}{\roles}\\
\text{Restricted concept expressions:} & \qquad \self{\roles} \qquad  \atleast{n}\roles.\concepts \qquad  \atmost{n}\roles.\concepts.
\end{align*}

The other structural restriction that is relevant for $\SROIQ$ is called \emph{regularity} and is concerned with RBox axioms only. Roughly speaking, the restriction ensures that cyclic dependencies between complex role inclusion axioms occur only in a limited form. For details, please see the pointers given in Section~\ref{sec:reading}.
%
For the introductory treatment in this paper, it suffices to note that regularity, just like simplicity, is a property of the ontology as a whole that cannot be checked for each axiom individually. An important practical consequence is that the union of two regular ontologies may no longer be regular. This must be taken into account when merging ontologies in practice.

\section{Description Logic Semantics}\label{sec:semantics}

The formal meaning of DL axioms is given by their model-theoretic semantics. In particular, the semantics specifies what the logical consequences of an ontology are. The formal semantics is therefore the main guideline for every tool that computes logical consequences of DL ontologies, and a basic understanding of its working is vital to make reasonable modelling choices and to comprehend the results given by software applications. Luckily, the semantics of description logics is not difficult to understand provided that some common misconceptions are avoided.

Intuitively speaking, an ontology describes a particular situation in a given domain of discourse. For example, the axioms in Sections~\ref{sec:basic} and \ref{sec:construct} describe a particular situation in the ``families and relationships'' domain. However, ontologies usually cannot fully specify the situation that they describe. On the one hand, there is no formal relationship between the symbols we use and the objects that they represent: the individual name $\individual{julia}$, for example, is just a syntactic identifier with no intrinsic meaning. Indeed, the intended meaning of the identifiers in our ontologies has no influence on their formal semantics: what we know about them stems only from the ontological axioms. On the other hand, the axioms in an ontology typically do not provide complete information. For example, \eqref{julia:neq:john} and \eqref{john:eq:johnny} in Section~\ref{subsection:abox} state that some individuals are equal and that others are unequal, but in many other cases this information might be left unspecified.

Description logics have been designed to deal with such incomplete information. Rather than making default assumptions in order to fully specify one particular interpretation for each ontology, the DL semantics generally considers all the possible situations (i.e., states of the world) where the axioms of an ontology would hold (we also say: where the axioms are \emph{satisfied}). This characteristic is sometimes called the \emph{Open World Assumption} since it keeps unspecified information open.\footnote{A \emph{Closed World Assumption} ``closes'' the interpretation by
assuming that every fact not explicitly stated to be true is actually false.
Both terms are not formally specified and rather outline the general flavour of a semantics than any particular definition.} A logical consequence of an ontology is an axiom that holds in all interpretations that satisfy the ontology, i.e., something that is true in all conceivable states of the world that agree with what is said in the ontology. The more axioms an ontology contains, the more specific are the constraints that it imposes on possible interpretations, and the fewer interpretations exist that satisfy all of the axioms.
Conversely, if fewer interpretations satisfy an ontology, then more axioms hold in all of them, and more logical consequences follow from the ontology.
The previous two sentences imply that the semantics of description logics is \emph{monotonic}: additional axioms always lead to additional consequences, or, more informally, the more knowledge we feed into a DL system the more results it returns.

An extreme case is when an ontology is not satisfied in any interpretation. The ontology is then called \emph{unsatisfiable} or \emph{inconsistent}. In this case \emph{every} axiom holds vacuously in all of the (zero) interpretations that satisfy the ontology.
Such an ontology is clearly of no utility, and avoiding inconsistency (and checking for it in the first place) is therefore an important task during modelling.

\begin{table}[t]
\caption{Syntax and semantics of $\SROIQ$ constructors}
\label{table:constructors}
\centering\rowcolors[]{1}{white}{gray!10}
\begin{tabular}{lcc}
& Syntax & Semantics\\
\hline
\quad\emph{Individuals:} &  &\\
individual name & $a$ & $a^\I$ \\
\hline
\quad\emph{Roles:} &  &\\
atomic role & $R$ & $R^\I$\\
inverse role & $R^-$ & $\set{\tuple{x,y} \mid \tuple{y,x} \in R^\I}$\\
universal role & $U$ & $\Delta^\I \times \Delta^\I$\\
\hline
\quad\emph{Concepts:} &  & \\
atomic concept & $A$ & $A^\I$ \\
intersection & $C \sqcap D$ & $C^\I \cap D^\I$ \\
union & $C \sqcup D$  & $C^\I \cup D^\I$ \\
complement & $\neg C$ & $\Delta^\I \setminus C^\I$ \\
top concept & $\top$ & $\Delta^\I$ \\
bottom concept & $\bot$ & $\emptyset$ \\
existential restriction & $\exists R.C$ &
$\set{x \mid \text{some $R^\I$-successor of $x$ is in $C^\I$}}$\\
universal restriction & $\forall R.C$ &
$\set{x \mid \text{all $R^\I$-successors of $x$ are in $C^\I$}}$ \\
at-least restriction & $\atleast{n}R.C$ &
$\set{x \mid \text{at least $n$ $R^\I$-successors of $x$ are in $C^\I$}}$ \\
at-most restriction & $\atmost{n}R.C$ &
$\set{x \mid \text{at most $n$ $R^\I$-successors of $x$ are in $C^\I$}}$ \\
local reflexivity
& $\self{R}$ & $\set{x \mid \tuple{x, x} \in R^\I}$ \\
nominal & $\set{a}$ & $\set{a^\I}$\\
\hline
\multicolumn{3}{l}{where $a,b\in\indnames$ are individual names, $A\in\connames$ is a concept name, $C,D\in\concepts$ are concepts, $R\in\roles$ is a role}\\
\hline
\end{tabular}
\end{table}
We have outlined above the most important ideas of DL semantics. What remains to be done is to define what we really mean by an ``interpretation'' and which conditions must hold for particular axioms to be satisfied by an interpretation. For this, we closely follow the intuitive ideas established above: an interpretation $\I$ consists of a set $\Delta^\I$ called the \emph{domain} of $\I$ and an interpretation function $\cdot^\I$ that maps each atomic concept $A$ to a set $A^\I\subseteq\Delta^\I$, each atomic role $R$ to a binary relation $R^\I\subseteq\Delta^\I\times\Delta^\I$, and each individual name $a$ to an element $a^\I\in\Delta^\I$. The interpretation of complex concepts and roles follows from the interpretation of the basic entities. Table~\ref{table:constructors} shows how to obtain the semantics of each compound expression from the semantics of its parts. By ``$R^\I$-successor of $x$'' we mean any individual $y$ such that $\tuple{x,y}\in R^\I$. The definition should confirm the intuitive explanations given for each case in Section~\ref{sec:construct}. For example, the semantics of $\concept{Female} \sqcap \concept{Parent}$ is indeed the intersection of the semantics of $\concept{Female}$ and $\concept{Parent}$.

\begin{table}[t]
\caption{Syntax and semantics of $\SROIQ$ axioms}
\label{table:axioms}
\centering\rowcolors[]{1}{white}{gray!10}
\begin{tabular}{lcc}
& Syntax & Semantics\\
\hline
\quad\emph{ABox:} &&\\
concept assertion & $C(a)$ & $a^\I \in C^\I$ \\
role assertion & $R(a, b)$ & $\tuple{a^\I, b^\I} \in R^\I$ \\
individual equality & $a \approx b$ & $a^\I = b^\I$ \\
individual inequality & $a \not\approx b$ & $a^\I \neq b^\I$ \\
\hline
\quad\emph{TBox:} &&\\
concept inclusion & $C \sqsubseteq D$ & $C^\I \subseteq D^\I$ \\
concept equivalence & $C \equiv D$ & $C^\I = D^\I$ \\
\hline
\quad\emph{RBox:} &&\\
role inclusion & $R \sqsubseteq S$ & $R^\I \subseteq S^\I$ \\
role equivalence & $R \equiv S$ & $R^\I = S^\I$ \\
complex role inclusion & $R_1 \circ R_2 \sqsubseteq S$ &
$R_1^\I \circ R_2^\I \subseteq S^\I$ \\
role disjointness & $\disjoint{R}{S}$ & $R^\I \cap S^\I
= \emptyset$ \\
\hline
\end{tabular}
\end{table}
Since an interpretation $\I$ fixes the meaning of all entities, we can unambiguously say for each axiom whether it holds in $\I$ or not. An axiom $\alpha$ \emph{holds} in $\I$ (we also say $\I$ \emph{satisfies} $\axiom$ and write $\I\models\axiom$) if the corresponding condition in Table~\ref{table:axioms} is met. Again, these definitions fully agree with the intuitive explanations given in Section~\ref{sec:basic}. If all axioms in an ontology $\ontology$ hold in $\I$ (i.e., if $\I$ satisfies $\ontology$, written $\I\models\ontology$), then $\I$ is a \emph{model} of $\ontology$. Thus a model is an abstraction of a state of the world that satisfies all axioms in the ontology. An ontology is \emph{consistent} if it has at least one model. An axiom $\axiom$ is a
\emph{consequence} of an ontology $\ontology$ (or $\ontology$ \emph{entails} $\alpha$, written $\ontology\models\alpha$) if $\alpha$ holds in every model of $\ontology$. In particular, an inconsistent ontology entails every axiom.

A noteworthy consequence of this semantics is the meaning of individual names in DL ontologies. We already remarked that DLs do not usually make the Unique Name Assumption, and indeed our formal definition allows two individual names to be interpreted as the same individual (element of the domain). Possibly even more important is the fact that the domain of an interpretation is allowed to contain many individuals that are not represented by any individual name. A common confusion in modelling arises from the implicit assumption that interpretations must only contain individuals that are represented by individual names (such individuals are also called \emph{named individuals}). For example, one could wrongly assume the ontology consisting of the axioms
\begin{equation*}
\role{parentOf}(\individual{julia},\individual{john})\qquad\concept{manyChildren}(\individual{julia})\qquad \concept{manyChildren}\sqsubseteq\atleast{3}\role{parentOf}.\top
\end{equation*}
 to be inconsistent since it requires Julia to have at least 3 children when only one (John) is given. However, there are many conceivable models where Julia does have three children, even though only one of the children is explicitly named. A significant number of modelling errors can be traced back to similar misconceptions that are easy to prevent if the general open world assumption of DLs is kept in mind.

Another point to note is that the above specification of the semantics does not provide any hint as to how to compute the relevant entailments in practical software tools.
There are infinitely many possible interpretations, each of which may have an infinite domain (in fact there are some ontologies that are satisfied only by interpretations with infinite domains). Therefore it is impossible to test all interpretations to see if they model a given ontology, and impossible to test all models of an ontology to see if they entail a given axiom.
Rather, one has to devise deduction procedures and prove their correctness with respect to the above specification. The interplay of certain expressive features can make reasoning algorithms more complicated and in some cases it can even be shown that no correct and terminating algorithm exists at all (i.e., that reasoning is undecidable). For our purposes it suffices to know that entailment of axioms is decidable for $\SROIQ$ (with the structural restrictions explained in Section~\ref{sec:syntax}) and that a number of free and commercial tools are available. Such tools are typically optimised for more specific reasoning problems, such as consistency checking, the entailment of concept subsumptions (subsumption checking) or of concept assertions (instance checking). Many of these standard inferencing problems can be expressed in terms of each other, so they can be handled by very similar reasoning algorithms.

\section{Important Fragments of $\SROIQ$}\label{sec:fragments}

Many different description logics have been introduced in the literature. Typically, they can be characterised by the types of constructors and axioms that they allow, which are often a subset of the constructors in $\SROIQ$. For example, the description logic $\mathcal{ALC}$ is the fragment of $\SROIQ$ that allows no RBox axioms and only $\sqcap$, $\sqcup$, $\neg$, $\exists$ and $\forall$ as its concept constructors. 
The extension of $\mathcal{ALC}$ with transitive roles is traditionally denoted by the letter $\mathcal{S}$.
Some other letters used in DL names hint at a particular constructor, such as inverse roles $\mathcal{I}$, nominals $\mathcal{O}$, qualified number restrictions $\mathcal{Q}$, and role hierarchies (role inclusion axioms without composition) $\mathcal{H}$.
So, for example, the DL named $\mathcal{ALCHIQ}$ extends $\mathcal{ALC}$ with role hierarchies, inverse roles and qualified number restrictions.
The letter $\mathcal{R}$ most commonly refers to the presence of role inclusions, local reflexivity $\Self$, and the universal role $U$, as well as the additional role characteristics of transitivity, symmetry, asymmetry, role disjointness, reflexivity, and irreflexivity. This naming scheme explains the name $\SROIQ$.

In recent years, fragments of DLs have been specifically developed in order to obtain favourable computational properties. For this purpose, $\mathcal{ALC}$ is already too large, since it only admits reasoning algorithms that run in worst-case exponential time. More lightweight DLs can be obtained by further restricting expressivity, while at the same time a number of additional $\SROIQ$ features can be added without loosing the good computational properties. The three main approaches for obtaining lightweight DLs are $\mathcal{EL}$, \emph{DLP} and \emph{DL-Lite}, which also correspond to language fragments OWL~EL, OWL~RL and OWL~QL of the Web Ontology Language.

The $\mathcal{EL}$ family of description logics is characterised by allowing unlimited use of existential quantifiers and concept intersection. The original description logic $\mathcal{EL}$ allows only those features and $\top$ but no unions, complements or universal quantifiers, and no RBox axioms. Further extensions of this language are known as $\mathcal{EL}^+$ and $\mathcal{EL}^{++}$. The largest such extension allows the constructors $\sqcap$, $\top$, $\bot$, $\exists$, $\Self$, nominals and the universal role, and it supports all types of axioms other than role symmetry, asymmetry and irreflexivity.
Interestingly, all standard reasoning tasks for this DL can still be solved in worst-case polynomial time. One can even drop the structural restriction of regularity that is important for $\SROIQ$. $\mathcal{EL}$ has been used to model large but lightweight ontologies that consist mainly of terminological data, in particular in the life sciences. A number of reasoners are specifically optimised for handling $\mathcal{EL}$-type ontologies, the most recent of which is the ELK reasoner for OWL~EL.\footnote{\url{http://elk-reasoner.googlecode.com/}}

DLP is short for \emph{Description Logic Programs} and comprises various DLs that are syntactically restricted in such a way that axioms could also be read as rules in first-order Horn logic without function symbols. Due to this, DLP-type logics can be considered as kinds of rule languages (hence the name OWL~RL) contained in DLs. To accomplish this, one has to allow different syntactic forms for subconcepts and superconcepts in concept inclusion axioms. We do not provide the details here.
While DLs in general may require us to consider domain elements that are not represented by individual names, for DLP one can always restrict attention to
models in which all domain elements are represented by individual names.
This is why DLP is often used to augment databases (interpreted as sets of ABox axioms), e.g., in an implementation of OWL~RL in the Oracle~11g database management system.

DL-Lite is a family of DLs that is also used in combination with large data collections and existing databases, in particular to augment the expressivity of a query language that retrieves such data.
This approach, known as Ontology Based Data Access, considers ontologies as a language for constructing \emph{views} or \emph{mapping rules} on top of existing data. The core feature of DL-Lite is that data access can be realised with standard query languages such as SQL that are not aware of the DL semantics. Ontological information is merely used in a query preprocessing step. Like DLP, DL-Lite requires different syntactic restrictions for subconcepts and superconcepts. We do not present the details here.

\section{Relationship to OWL}\label{sec:owl}

The \emph{OWL Web Ontology Language} is a knowledge representation language standardised by the World Wide Web Consortium (W3C). OWL is one of the most important applications of description logics today. In this section, we briefly outline the relationship of the two languages. A comprehensive treatment is beyond the scope of this paper; see Section~\ref{sec:reading} for pointers to further reading. The current version of the OWL specification is OWL~2 as standardised in 2009. This supersedes the earlier OWL~1 standard of 2004.

The main building blocks of OWL are indeed very similar to those of DLs, with the main difference that concepts are called \emph{classes} and roles are called \emph{properties}. It is therefore not surprising that description logics have had a major influence on the development of OWL and the expressive features that it provides. Historically, however, OWL has also been conceived as an extension to RDF, a Web data modelling language whose expressivity is comparable to DL ABoxes. The formal semantics of RDF is subtly different from that of DLs, even though both lead to the same consequences in many common cases. Extending the RDF semantics to the expressive features of OWL improves the compatibility between the two, but it also makes reasoning undecidable. Therefore, it has been decided to specify both styles of formal semantics for OWL: the \emph{Direct Semantics} based on DLs and the \emph{RDF-based Semantics}.

In this section, we are therefore mainly interested in the Direct Semantics of OWL. This semantics is only defined for OWL ontologies that abide by certain syntactic restrictions (essentially the restriction that the OWL axioms can be read as $\SROIQ$ axioms for which the structural restrictions of Section~\ref{sec:syntax} are satisfied). This syntactic fragment of OWL is called \emph{OWL~DL}.\footnote{In contrast, the OWL language without any syntactic constraints is called \emph{OWL~Full}. It comprises ontologies that can only be interpreted under the RDF-based Semantics.} Under the Direct Semantics, large parts of OWL~DL can indeed be considered as a syntactic variant of $\SROIQ$. For example, the axiom $\concept{Mother} \equiv \concept{Female} \sqcap \concept{Parent}$ would be written as follows in OWL:
\begin{equation*}
\texttt{EquivalentClasses}(~\concept{Mother}~~\texttt{ObjectIntersectionOf}(~\concept{Female}~~\concept{Parent}~)~)
\end{equation*}
where the symbols $\concept{Mother}$, $\concept{Female}$ and $\concept{Parent}$ would be identifier strings that conform to the OWL specification.\footnote{Entity names in OWL are generally based on Uniform Resource Identifiers (URIs). The details are not relevant here.} The above example illustrates the close relationship between the syntax of $\SROIQ$ and that of OWL. In many cases, it is indeed enough to translate an operator symbol of $\SROIQ$ into the corresponding operator name in OWL, which is then written in prefix notation like a function. This is also why the above form of syntax is called \emph{Functional-Style Syntax}. The OWL standard provides a number of syntactic forms that can be used to express OWL ontologies. The most prominent among these is the RDF/XML serialisation since it is the only format that all conforming OWL tools need to understand. On the other hand, it is more difficult for humans to read and we do not present it here.

It is interesting to note that there are still a few differences between OWL~DL under the Direct Semantics and $\SROIQ$. On a syntactic level, OWL provides a lot more operators that, though logically redundant, can be convenient as shortcuts for compound DL axioms. For example, OWL has special constructs for specifying domain and range of a property, even though these could equally well be expressed as in Section~\ref{subsection:role:restrictions}. These kinds of features also include the empty (bottom) property, which can easily be defined but is not included as a language feature in DLs.

However, OWL also includes some expressive features that we did not include in our treatment of $\SROIQ$ above. Most notably, this includes support for datatypes and datatype literals. These behave like classes and individual names but come with a fixed, pre-defined interpretation. For example, the datatype for Boolean values has exactly two elements -- true and false -- in any interpretation. This can also be introduced in DLs by so-called \emph{concrete domains}, i.e., pre-defined interpretation domains. Both DLs and OWL in this case strictly distinguish roles/properties that relate to ``abstract'' individuals from those that relate to values from some datatype. In OWL, the constructs that relate to datatypes include ``Data'' in their name while constructs that relate to abstract individuals include ``Object.'' For example, OWL distinguishes \texttt{ObjectIntersectionOf} (used above) from \texttt{DataIntersectionOf} (the intersection of datatypes).

The only other logical feature that is missing in DLs are so-called \emph{Keys}. These are special forms of rules that can be used for data integration. Roughly speaking, a key specifies that two named individuals are entailed to be equal if they agree on certain property values and class memberships, similar to key constraints in databases.
For example, the combination of nationality and registration number might be treated as a key for (i.e., sufficient to uniquely identify) motor vehicles.

Besides the logical features, OWL also includes a number of other aspects that are not considered in description logics at all. For example, it includes means of naming an ontology and of importing ontological axioms from one ontology into another.
Further extra-logical features include a simple form of \emph{meta-modelling} called \emph{punning}, non-logical axioms to \emph{declare} identifiers, and the possibility to add \emph{annotations} to arbitrary axioms and entities similar to comments in a programming language.
%

\section{Further Reading}\label{sec:reading}

This paper can only provide a first introduction to description logics and OWL.
More detailed introductory texts can be found in the lecture notes of the Reasoning Web Summer School:
Rudolph provides a detailed discussion of DL semantics and modelling \cite{rudolph2011fodl},
Baader gives a general overview with extended historical notes \cite{baader2009rwdl}, and
Sattler focusses on tableau-based reasoning methods \cite{sattler2007rwdl}.
An extensive introduction to lightweight description logics is given by Kr\"otzsch \cite{kroetzsch2012lightweightDLs}.

For a more detailed coverage of OWL and its relationship to DL, we recommend the textbook
\href{http://www.semantic-web-book.org/}{\emph{Foundations of Semantic Web Technologies}} \cite{fost}.
This introductory text also treats the relationship of DLs to first-order logic, DL query answering and extensions for rule-based modelling
(related to keys in OWL), which we have omitted here. An in-depth treatment of description logics and related research topics
is provided by the \emph{Description Logic Handbook} \cite{dlhandbook}, which also covers interesting aspects of deduction
algorithms and computational complexity that are beyond the scope of this paper.

A number of research papers focus on specific topics in DLs. Closely related to this paper is the original article on $\SROIQ$, which also provides the
details on regularity conditions that have been skipped above \cite{HKS06-sroiq}.
A detailed discussion of OWL datatypes and their description logic semantics is given by Motik and Horrocks \cite{MH08:owldatatypes}.
There are also various works that focus on $\mathcal{EL}$ \cite{BBL-EL,Kroetzsch11:elreason}, DLP/OWL~RL \cite{DLP-www03,Kroetzsch12:owlrl} and DL-Lite \cite{dl-lite}.
Current developments in DL research are discussed at the annual DL Workshop (see \url{http://dl.kr.org/} for proceedings) and at the major Semantic Web and Artificial Intelligence conferences.

The primary resources on OWL~2 are the online documents of the specification \cite{owl2-overview} where the OWL Primer provides a first introduction \cite{owl2-primer}. The differences of the 2009 OWL~2 standard to its predecessor are explained in \cite{CG+:OWLtwo08}.

Many related tools such as reasoners and ontology editors are available. The most popular free ontology editor is Prot\'{e}g\'{e},\footnote{\url{http://protege.stanford.edu/}} which can be used with a variety of OWL reasoners. Pointers to current OWL reasoners are best found online.\footnote{A list of reasoners can be found, e.g., at \url{http://semanticweb.org/wiki/Category:Reasoner}.} Popular systems for large parts of OWL~2~DL ($\SROIQ$) include FaCT++, HermiT, Pellet and RacerPro. Some typical lightweight systems are ELK (OWL~EL), jCEL (OWL~EL), Owlgress (OWL~QL), OWLIM (OWL~RL and QL), Quonto (OWL~QL) and Snorocket (OWL~EL). Details about these tools and related publications can be found on the respective homepages.

\paragraph*{Acknowledgements}
We thank Fernando Bobillo, Peter Patel-Schneider and Evgeny Zolin for helpful comments on an earlier version of this text.

\bibliographystyle{plain}
\bibliography{references}

\begin{thebibliography}{10}

\bibitem{baader2009rwdl}
Franz Baader.
\newblock Description logics.
\newblock In Sergio Tessaris, Enrico Franconi, Thomas Eiter, Claudio Gutierrez,
  Siegfried Handschuh, Marie-Christine Rousset, and Renate~A. Schmidt, editors,
  {\em Reasoning Web. Semantic Technologies for Information Systems -- 5th
  International Summer School, 2009}, volume 5689 of {\em LNCS}, pages 1--39.
  Springer, 2009.
\newblock Available at \url{http://lat.inf.tu-dresden.de/research/papers.html}.

\bibitem{BBL-EL}
Franz Baader, Sebastian Brandt, and Carsten Lutz.
\newblock Pushing the $\mathcal{EL}$ envelope.
\newblock In {Leslie Pack} Kaelbling and Alessandro Saffiotti, editors, {\em
  Proc.\ 19th Int.\ Joint Conf.\ on Artificial Intelligence (IJCAI'05)}, pages
  364--369. Professional Book Center, 2005.

\bibitem{dlhandbook}
Franz Baader, Diego Calvanese, Deborah McGuinness, Daniele Nardi, and Peter
  Patel-Schneider, editors.
\newblock {\em The Description Logic Handbook: Theory, Implementation, and
  Applications}.
\newblock Cambridge University Press, second edition, 2007.

\bibitem{dl-lite}
Diego Calvanese, Guiseppe~De Giacomo, Domenico Lembo, Maurizio Lenzerini, and
  Riccardo Rosati.
\newblock Tractable reasoning and efficient query answering in description
  logics: The {DL-Lite} family.
\newblock {\em J.\ of Automated Reasoning}, 39(3):385--429, 2007.

\bibitem{CG+:OWLtwo08}
Bernardo {Cuenca Grau}, Ian Horrocks, Boris Motik, Bijan Parsia, Peter
  Patel-Schneider, and Ulrike Sattler.
\newblock {OWL 2}: {The} next step for {OWL}.
\newblock {\em J.\ of Web Semantics}, 6:309--322, 2008.

\bibitem{DLP-www03}
Benjamin~N. Grosof, Ian Horrocks, Raphael Volz, and Stefan Decker.
\newblock Description logic programs: combining logic programs with description
  logic.
\newblock In {\em Proc.\ 12th Int.\ Conf.\ on World Wide Web (WWW'03)}, pages
  48--57. ACM, 2003.

\bibitem{owl2-primer}
Pascal Hitzler, Markus Kr\"o{}tzsch, Bijan Parsia, Peter~F. Patel-Schneider,
  and Sebastian Rudolph, editors.
\newblock {\em {OWL~2 Web Ontology Language: Primer}}.
\newblock W3C Recommendation, 27 October 2009.
\newblock Available at \url{http://www.w3.org/TR/owl2-primer/}.

\bibitem{fost}
Pascal Hitzler, Markus Kr\"{o}tzsch, and Sebastian Rudolph.
\newblock {\em Foundations of Semantic Web Technologies}.
\newblock Chapman \& Hall/CRC, 2009.

\bibitem{HKS06-sroiq}
Ian Horrocks, Oliver Kutz, and Ulrike Sattler.
\newblock The even more irresistible $\mathcal{SROIQ}$.
\newblock In Patrick Doherty, John Mylopoulos, and Christopher~A. Welty,
  editors, {\em Proc.\ 10th Int.\ Conf.\ on Principles of Knowledge
  Representation and Reasoning (KR'06)}, pages 57--67. AAAI Press, 2006.

\bibitem{Kroetzsch11:elreason}
Markus Kr{\"o}tzsch.
\newblock Efficient rule-based inferencing for {OWL EL}.
\newblock In Toby Walsh, editor, {\em Proc.\ 22nd Int.\ Conf.\ on Artificial
  Intelligence (IJCAI'11)}, pages 2668--2673. AAAI Press/IJCAI, 2011.

\bibitem{Kroetzsch12:owlrl}
Markus Kr{\"o}tzsch.
\newblock The not-so-easy task of computing class subsumptions in {OWL RL}.
\newblock In Philippe Cudr{\'e}-Mauroux, Jeff Heflin, Evren Sirin, Tania
  Tudorache, J{\'e}r{\^o}me Euzenat, Manfred Hauswirth, Josiane~Xavier
  Parreira, Jim Hendler, Guus Schreiber, Abraham Bernstein, and Eva Blomqvist,
  editors, {\em Proc.\ 11th Int.\ Semantic Web Conf.\ (ISWC'12)}, volume 7649
  of {\em LNCS}, pages 279--294. Springer, 2012.

\bibitem{kroetzsch2012lightweightDLs}
Markus Kr\"otzsch.
\newblock {OWL 2 Profiles}: An introduction to lightweight ontology languages.
\newblock In Thomas Eiter and Thomas Krennwallner, editors, {\em Proceedings of
  the 8th Reasoning Web Summer School, Vienna, Austria, September 3--8 2012},
  volume 7487 of {\em LNCS}, pages 112--183. Springer, 2012.
\newblock Available at \url{http://korrekt.org/page/OWL_2_Profiles}.

\bibitem{MH08:owldatatypes}
Boris Motik and Ian Horrocks.
\newblock {OWL} datatypes: {Design} and implementation.
\newblock In Amit Sheth, Steffen Staab, Mike Dean, Massimo Paolucci, Diana
  Maynard, Timothy Finin, and Krishnaprasad Thirunarayan, editors, {\em Proc.\
  7th Int.\ Semantic Web Conf.\ (ISWC'08)}, volume 5318 of {\em LNCS}, pages
  307--322. Springer, 2008.

\bibitem{owl2-overview}
W3C {OWL Working Group}.
\newblock {\em {OWL~2 Web Ontology Language: Document Overview}}.
\newblock W3C Recommendation, 27 October 2009.
\newblock Available at \url{http://www.w3.org/TR/owl2-overview/}.

\bibitem{rudolph2011fodl}
Sebastian Rudolph.
\newblock Foundations of description logics.
\newblock In Axel Polleres, Claudia d'Amato, Marcelo Arenas, Siegfried
  Handschuh, Paula Kroner, Sascha Ossowski, and Peter~F. Patel-Schneider,
  editors, {\em Reasoning Web. Semantic Technologies for the Web of Data -- 7th
  International Summer School 2011}, volume 6848 of {\em LNCS}, pages 76--136.
  Springer, 2011.
\newblock Available at \url{http://www.aifb.kit.edu/web/Incollection3026/en}.

\bibitem{RKH-elephants-08}
Sebastian Rudolph, Markus Kr\"{o}tzsch, and Pascal Hitzler.
\newblock All elephants are bigger than all mice.
\newblock In Franz Baader, Carsten Lutz, and Boris Motik, editors, {\em Proc.\
  21st Int.\ Workshop on Description Logics (DL'08)}, volume 353 of {\em CEUR
  Workshop Proceedings}. CEUR-WS.org, 2008.

\bibitem{RKH:Jelia-08}
Sebastian Rudolph, Markus Kr\"{o}tzsch, and Pascal Hitzler.
\newblock Cheap {Boolean} role constructors for description logics.
\newblock In Steffen H\"{o}lldobler, Carsten Lutz, and Heinrich Wansing,
  editors, {\em Proc.\ 11th European Conf.\ on Logics in Artificial
  Intelligence (JELIA'08)}, volume 5293 of {\em LNAI}, pages 362--374.
  Springer, 2008.

\bibitem{sattler2007rwdl}
Ulrike Sattler.
\newblock Reasoning in description logics: Basics, extensions, and relatives.
\newblock In Grigoris Antoniou, Uwe A\ss{}mann, Cristina Baroglio, Stefan
  Decker, Nicola Henze, Paula-Lavinia Patranjan, and Robert Tolksdorf, editors,
  {\em Reasoning Web -- 3rd International Summer School, 2007}, volume 4636 of
  {\em LNCS}, pages 154--182. Springer, 2007.

\end{thebibliography}

\end{document}